\documentclass[sigconf]{acmart}
\AtBeginDocument{%
  }

\setcopyright{acmlicensed}
\copyrightyear{2025}
\acmYear{2025}
\setcopyright{acmlicensed}\acmConference[RecSys '25]{Proceedings of the
Nineteenth ACM Conference on Recommender Systems}{September 22--26,
2025}{Prague, Czech Republic}
\acmBooktitle{Proceedings of the Nineteenth ACM Conference on
Recommender Systems (RecSys '25), September 22--26, 2025, Prague, Czech
Republic}
\acmDOI{10.1145/3705328.3748056}
\acmISBN{979-8-4007-1364-4/2025/09}

\usepackage{xcolor}
\definecolor{dullorange}{RGB}{204,119,34}

\usepackage{geometry}
\usepackage{tabularx}
\usepackage{booktabs}  
\usepackage{xcolor}    
\usepackage{array}     
\usepackage{lipsum}    
\usepackage{fancyvrb}
\usepackage{fvextra} 
\usepackage{calc}    
\usepackage{graphicx}
\usepackage{longtable}
\usepackage{multirow}

\begin{document}

\title{GRACE: Generative Recommendation via Journey-Aware Sparse Attention on Chain-of-Thought Tokenization}

\author{Luyi Ma}
\authornote{Equally contributed.}
\affiliation{%
  \institution{Walmart Global Tech}
  \city{Sunnyvale}
  \state{California}
  \country{USA}}
\email{luyi.ma@walmart.com}
\orcid{https://orcid.org/0009-0002-7454-3959}

\author{Wanjia Sherry Zhang}
\authornotemark[1]
\affiliation{%
  \institution{Walmart Global Tech}
  \city{Sunnyvale}
  \state{California}
  \country{USA}}
\email{sherry.zhang@walmart.com}
\orcid{https://orcid.org/0000-0002-9124-2843}

\author{Kai Zhao}
\authornotemark[1]
\affiliation{%
  \institution{Walmart Global Tech}
  \city{Sunnyvale}
  \state{California}
  \country{USA}}
\email{kai.zhao@walmart.com}
\orcid{https://orcid.org/0000-0003-1040-0211}

\author{Abhishek Kulkarni}
\authornotemark[1]
\affiliation{%
  \institution{Walmart Global Tech}
  \city{Bangalore}
  \state{Karnataka}
  \country{India}}
\email{abhishek.kulkarni@walmart.com}
\orcid{https://orcid.org/0009-0004-0698-2727}

\author{Lalitesh Morishetti}
\affiliation{%
  \institution{Walmart Global Tech}
  \city{Sunnyvale}
  \state{California}
  \country{USA}}
\email{lalitesh.morishetti@walmart.com}
\orcid{https://orcid.org/0009-0004-9085-9587}

\author{Anjana Ganesh}
\affiliation{%
  \institution{Walmart Global Tech}
  \city{Sunnyvale}
  \state{California}
  \country{USA}}
\email{anjana.ganesh@walmart.com}
\orcid{https://orcid.org/0009-0005-1164-7760}

\author{Ashish Ranjan}
\affiliation{%
  \institution{Walmart Global Tech}
  \city{Sunnyvale}
  \state{California}
  \country{USA}}
\email{ashish.ranjan0@walmart.com}
\orcid{https://orcid.org/0009-0009-6977-2387}

\author{Aashika Padmanabhan}
\affiliation{%
  \institution{Walmart Global Tech}
  \city{Sunnyvale}
  \state{California}
  \country{USA}}
\email{aashika.padmanabhan@walmart.com}
\orcid{https://orcid.org/0009-0007-2288-3851}

\author{Jianpeng Xu}
\affiliation{%
  \institution{Walmart Global Tech}
  \city{Sunnyvale}
  \state{California}
  \country{USA}}
\email{jianpeng.xu@walmart.com}
\orcid{https://orcid.org/0000-0003-3702-528X}

\author{Jason Cho}
\affiliation{%
  \institution{Walmart Global Tech}
  \city{Sunnyvale}
  \state{California}
  \country{USA}}
\email{jason.cho@walmart.com}
\orcid{https://orcid.org/0000-0002-8106-2961}

\author{Praveen Kanumala}
\affiliation{%
  \institution{Walmart Global Tech}
  \city{Sunnyvale}
  \state{California}
  \country{USA}}
\email{pkanumala@walmart.com}
\orcid{https://orcid.org/0009-0009-1606-8409}

\author{Kaushiki Nag}
\affiliation{%
  \institution{Walmart Global Tech}
  \city{Sunnyvale}
  \state{California}
  \country{USA}}
\email{kaushiki.nag@walmart.com}
\orcid{https://orcid.org/0009-0008-4859-2937}

\author{Sumit Dutta}
\affiliation{%
  \institution{Walmart Global Tech}
  \city{Bangalore}
  \state{Karnataka}
  \country{India}}
\email{sumit.dutta@walmart.com}
\orcid{https://orcid.org/0009-0004-9269-7198}

\author{Kamiya Motwani}
\affiliation{%
  \institution{Walmart Global Tech}
  \city{Bangalore}
  \state{Karnataka}
  \country{India}}
\email{kamiya.motwani@walmart.com}
\orcid{https://orcid.org/0009-0004-1030-3299}

\author{Malay Patel}
\affiliation{%
  \institution{Walmart Global Tech}
  \city{Sunnyvale}
  \state{California}
  \country{USA}}
\email{mpatel@walmart.com}
\orcid{https://orcid.org/0009-0009-4819-5090}

\author{Evren Korpeoglu}
\affiliation{%
  \institution{Walmart Global Tech}
  \city{Sunnyvale}
  \state{California}
  \country{USA}}
\email{EKorpeoglu@walmart.com}
\orcid{https://orcid.org/0009-0003-7754-3652}

\author{Sushant Kumar}
\affiliation{%
  \institution{Walmart Global Tech}
  \city{Sunnyvale}
  \state{California}
  \country{USA}}
\email{sushant.kumar@walmart.com}
\orcid{https://orcid.org/0009-0000-5643-5263}

\author{Kannan Achan}
\affiliation{%
  \institution{Walmart Global Tech}
  \city{Sunnyvale}
  \state{California}
  \country{USA}}
\email{kannan.achan@walmart.com}
\orcid{https://orcid.org/0009-0000-9186-3175}

\renewcommand{\shortauthors}{Luyi Ma et al.}

\begin{abstract}
Generative models have recently demonstrated strong potential in multi-behavior recommendation systems, leveraging the expressive power of transformers and tokenization to generate personalized item sequences.
However, their adoption is hindered by (1) the lack of explicit information for token reasoning, (2) high computational costs due to quadratic attention complexity and dense sequence representations after tokenization, and (3) limited multi-scale modeling over user history. In this work, we propose GRACE (Generative Recommendation via journey-aware sparse Attention on Chain-of-thought tokEnization), a novel generative framework for multi-behavior sequential recommendation. GRACE introduces a hybrid Chain-of-Thought (CoT) tokenization method that encodes user-item interactions with explicit attributes from product knowledge graphs (e.g., category, brand, price) over semantic tokenization, enabling interpretable and behavior-aligned generation. To address the inefficiency of standard attention, we design a Journey-Aware Sparse Attention (JSA) mechanism, which selectively attends to compressed, intra-, inter-, and current-context segments in the tokenized sequence. Experiments on two real-world datasets show that GRACE significantly outperforms state-of-the-art baselines, achieving up to +106.9\% HR@10 and +106.7\% NDCG@10 improvement over the state-of-the-art baseline on the Home domain, and +22.1\% HR@10 on the Electronics domain. GRACE also reduces attention computation by up to 48\% with long sequences. 
\end{abstract}




\maketitle

\section{Introduction}

\begin{figure}[t]
    \centering
    \includegraphics[width=\linewidth]{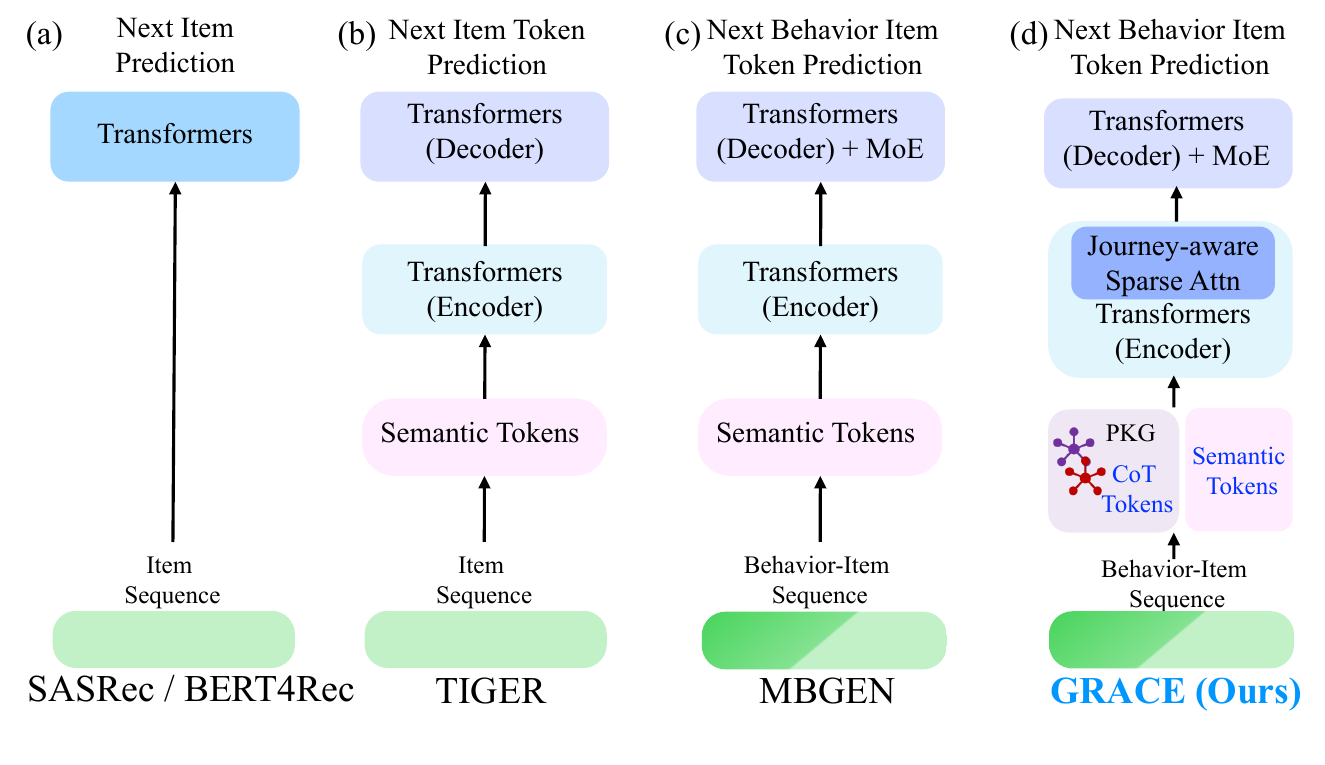}
    \caption{Comparison with existing generative recommendation model structures.}
    \label{figure1}
\end{figure}

The evolution of recommendation systems has been marked by a continuous pursuit of more accurate and context-aware user personalization, progressing from simple sequential models to increasingly sophisticated generative architectures. At the core of this transformation lies the transformer model (Figure~\ref{figure1}a), known for its powerful sequence modeling capabilities on item ID sequence through attention mechanisms~\cite{kang2018self, sun2019bert4rec, de2021transformers4rec, yao2024recommender}.
While effective in capturing sequential dependencies, these item ID-based models lack the capacity to represent meaningful item information during candidate generation. 

Recent works on generative recommendation address these limitations, and TIGER~\cite{rajput2023recommender} (Figure~\ref{figure1}b) introduced tokenization through RQ-VAE~\cite{zeghidour2021soundstream} based on item descriptive text to transform item ID sequences into coarse-grained to fine-grained semantic tokens, thereby improving recommendation context-awareness. However, TIGER's abstraction was largely limited to the item level, overlooking behavioral context, which plays a critical role in real-world recommendations.

The multi-behavior sequential recommendation (MBSR) problem predicts the items the user will interact with by incorporating behavior types of history user-item~\cite{yang2022multi, yang2022multi, su2023personalized} and MBGen~\cite{liu2024multi} (Figure~\ref{figure1}c) pushed the boundary further by incorporating behavior-item tokenization and extending the encoder-decoder framework for generative modeling with multi-behavior support. This allowed the model to learn richer context and generate token sequences aligned with behavioral signals. Nevertheless, MBGen relies on static tokenization and full attention, leading to increased memory and computation costs, especially with long and diverse interaction histories after tokenization. Moreover, the semantic tokens still lack deterministic information such as explicit item attributes in the product knowledge graph (PKG) due to the unsupervised objectives~\cite{rajput2023recommender, zeghidour2021soundstream}.

We answer two major research questions in this paper. \textbf{RQ1}: how to enhance explicit and deterministic information on top of semantic tokens for better model performance; \textbf{RQ2}: how to improve the full attention on tokenized user-item behavior sequence which is inefficient in computation and learning capacity, where a linear growth in token length results in a quadratic growth in full attention, and more tokens complicates the learning of multi-behavior patterns and multi-scale interests.

In this paper, we present GRACE, a novel architecture designed to address the aforementioned challenges hindering existing multi-behavior generative recommender systems. 
As illustrated in Figure~\ref{figure1}d, GRACE innovatively integrates two key components: a hybrid hierarchical tokenization framework and a Journey-aware Sparse Attention (JSA).
To answer \textbf{RQ1}, GRACE improves the semantic tokenization with a hybrid multi-level tokenization, incorporating implicit semantic features from the existing semantic tokens and explicit structured knowledge from a product knowledge graph (PKG)~\cite{zalmout2021all,chen2024relation} traversal through deterministic item attributes like category, brand, price. Such a coarse-grained-to-fine-grained PKG traversal forms a Chain-of-Thought~\cite{wei2022chain} (CoT) path, mimicking a user's reasoning or decision-making process from high-level product categories to fine-grained item details. This token structure enhances interpretability and the capability to predict next interactions with greater semantic alignment and behavioral awareness.

\begin{figure*}[t]
    \centering
    \includegraphics[width=0.95\textwidth]{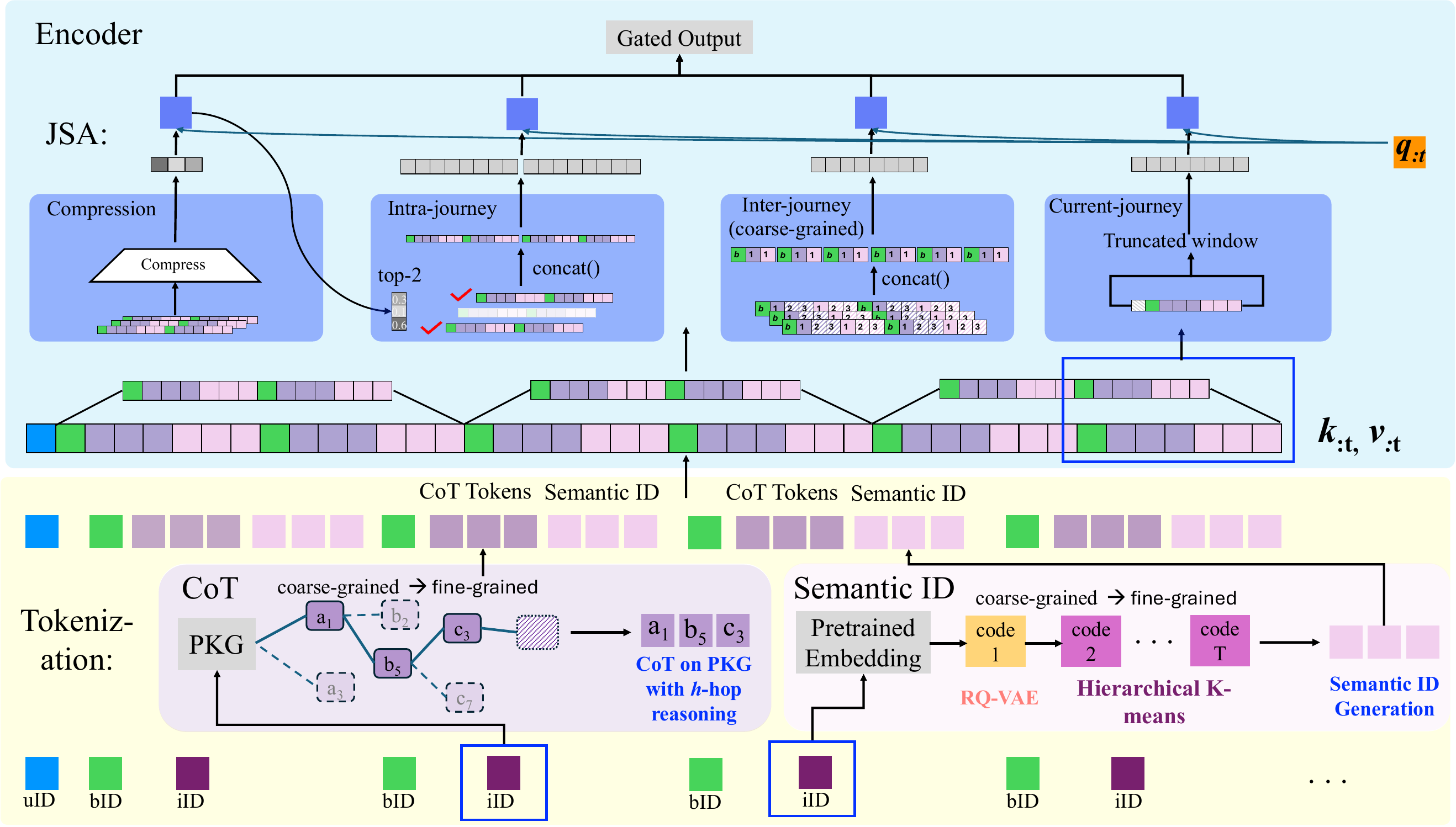}
    \caption{Illustration of GRACE framework with Hybrid Tokenization and Journey-aware Sparse Attention (JSA). Hybrid Tokenization (bottom) contains Chain-of-Thought tokens and semantic IDs. The JSA (top) is a combination of compression, intra-journey, inter-journey and current-journey attentions.
    }
    \label{figure2}
\end{figure*}

To address \textbf{RQ2} on the rich and long token sequence, GRACE introduces Journey-aware Sparse Attention (JSA)—a trainable sparse attention mechanism specifically tailored for multi-behavior recommendation.
JSA decomposes the attention process into four complementary scopes: (1) compression over segmented journey blocks, (2) selection of top-N intra-journey blocks for important customer interest, (3) inter-journey traversal through coarse-grained tokens for high-level journey transition, and (4) a truncated window for current-context modeling. This design allows GRACE to dynamically allocate attention to only the most relevant tokens across different temporal and semantic scopes, drastically reducing computational overhead while maintaining the expressiveness required for complex recommendation tasks. 
To evaluate CoT tokenization and JSA, we collect the real-world user behavior sequence and the product attributes. The extensive experiments on these real-world datasets demonstrate the superiority of our proposed method over state-of-the-art baselines.

Our main contributions with GRACE are three-fold: 
\begin{itemize}
    \item We propose the CoT tokenization framework in generative recommendations for MBSR that integrates semantic signals and CoT knowledge from product knowledge graphs (PKG), enabling interpretable and structured sequence generation for recommendation.
    \item We introduce Journey-aware Sparse Attention (JSA), a trainable sparse attention module that captures intra-journey continuity, inter-journey transitions, and current context, offering substantial improvements in efficiency and scalability.
    \item We conduct extensive experiments on real-world datasets, showing that GRACE outperforms existing methods in recommendation accuracy, training/inference efficiency, and behavior-level robustness, offering a practical solution for next-generation generative recommender systems.
\end{itemize}

\section{METHODOLOGY}

\subsection{Problem Definition}
We define our approach to the multi-behavior sequential recommendation problem by considering the historical interaction sequence and corresponding behavior information (e.g., add-to-cart, view, remove-from-cart, and like) to predict the next item to interact with. 
Formally, we denote the multi-behavior interaction sequence for a user $u \in \mathcal{U}$ as $s_u = [(v_1, b_1), ..., (v_{n-1}, b_{n-1})]$ with item $v \in \mathcal{V}$ and behavior information $b \in \mathcal{B}$. The last interaction $(v_n, b_n)$ of $u$ is the prediction target. Typically, we predict the target behavior $b_n$ first based on $s_u$ (\textbf{Stage 1}) by modeling the probability distribution $P(b_n | s_u)$, and then we predict the target item $v_n$ based on $s_u + [b_n]$ (\textbf{Stage 2}) by modeling $P(v_n | b_n, s_u)$. Combining two tasks leads to $P(b_n, v_n | s_u) = P(v_n| b_n, s_u) \cdot P(b_n | s_u)$ for next item interaction $(b_n, v_n)$ prediction.

\noindent \textbf{Recommendation Tasks}: To better model these two stages in MBSR, we adopt the setting of (1) \textit{target behavior item prediction task}, (2) \textit{behavior-specific item prediction task}, and (3) \textit{behavior-item prediction task} \cite{liu2024multi}. In target behavior item prediction, the behavior in Stage 1 is provided by a single behavior (e.g., Add-to-cart), and the model predicts the target item. In behavior-specific item prediction, the behavior in Stage 1 is provided by a set of predefined behaviors instead of a single behavior. In behavior-item prediction, the model is to predict both the behavior and the item sequentially without given information. 
Our model will take the tokenized $s_u$ as input and encode the information through the attention blocks with JSA (Figure \ref{figure2}). The decoder processes the encoder's output to complete three prediction tasks (Figure \ref{figure1}d and \ref{nsainfer}). 




\subsection{Overview}
GRACE builds on the encoder-decoder architecture similar to previous generative recommendation architecture (Figure \ref{figure1}d). The encoder module focuses on learning the rich contexts from item information for recommendation prediction, and the decoder module focuses on token prediction of the target behavior-item pair. This is because the target item $v_n$ is expanded to a token sequence after tokenization, and the prediction of $(b_n, v_n)$ becomes a sequence-to-sequence problem.
The encoder (decoder) module stacks multiple attention blocks to model sequence transformation, and each attention block could consist of an attention layer and a Mixture-of-Experts (MoE) layer. 
Formally, given the input sequence $X$, we define the multi-head attention as follows, where the queries, keys, and values are matrices $Q_i=W_i^QX, K_i=W_i^K X$ and $V_i=W_i^V X$: 
\begin{align}
        o_i &= \textbf{Attn}(Q_i, K_i, V_i), \text{ }i\in[1, H],
\end{align}
where $H$ is the number of heads. The MoE layer takes the output of the attention module and applies a routing mechanism over the output of multiple experts, which are usually implemented by an MLP.

\subsection{Chain-of-Thought Tokenization}
We tokenize both items and behaviors to keep richer information for modeling. We convert each behavior $b \in \mathcal{B}$ (`bID') into a unique token ({\color{black}green square} in Figure~\ref{figure2}). 
When a user shops for an item, she usually combines the information from all these modalities to decide the next item to interact with. 
We could tokenize each item $v \in \mathcal{V}$ by (1) semantic tokenization based on items' rich meta information, such as item title, description, and images for item identification, and (2) Chain-of-Thought (CoT) tokenization based on the attributes in PKG for more granular navigation.

\subsubsection{Semantic Tokenization}
An item $v$'s semantic information is critical to distinguish itself from other items. Given the meta information $\mathcal{M}$ of an item $v$, 
We convert the item meta information $v^{\mathcal{M}}$ to embeddings through pre-trained encoders, $\textbf{e}_v^{\mathcal{M}} = \mathcal{F}_{t}(v_{\mathcal{M}})$. We adopt the balanced semantic ID approach from MBGen~\cite{liu2024multi} by combining Residual-Quantized Variational Autoencoder (RQ-VAE)
and K-means clustering to generate $T$-level semantic tokens. The model predicts all $T$ semantic tokens $ST_v$ generatively to predict the next item (Eq. \ref{eq:h-kmeans}). Figure~\ref{figure2} Semantic ID summarizes this process. 
\begin{equation}
    ST_v = \textbf{Balanced-SemanticID}(\textbf{e}_v^{\mathcal{M}}, T) = [st_v^1, ..., st_v^T]
    \label{eq:h-kmeans}
\end{equation}

\subsubsection{Chain-of-Thought Tokenization}
Navigating through PKG from coarse-grained product categories to fine-grained product-to-purchase is usually the user's shopping journey to find the right items\footnote{https://hbr.org/2023/03/high-tech-touchpoints-are-changing-customer-experience.}. 
Typically, a PKG could return a list of \textit{h-hop} attributes from coarse-grained to fine-grained given the provided item $v$, $PKG[v] = [a_1, b_5, c_3, ...]$.
To incorporate the hierarchical information, we convert each unique attribute to a token and insert it between the behavior token and the semantic tokens. 
In the generative recommendation, tokenizing the nodes in PKG on the attribute trajectory toward an item forms a step-by-step reasoning, from coarse-grained to fine-grained, to predict the semantic tokens of an item. 
Such a Chain-of-Thought process on PKG could provide a richer and more deterministic context to model item transitions. 
As visualized in Figure \ref{figure2}, we combine the behavior tokens ({\color{black}green square}), the $T$-level semantic tokens ({\color{black}pink square}) and $h$-hop graph CoT tokens ({\color{black}purple square}), and convert a user $u$'s interaction sequence $s_u$ into token sequence, with a user token at the first position. The tokenized sequence will be the input for the encoder and the decoder for modeling.

\subsection{Journey-Aware Sparse Attention}
While tokenization of user interaction sequences preserves rich features, it also increases the sequence length linearly and the attention matrix size quadratically, which requires a more efficient attention mechanism to learn the item transition pattern. We introduce a unified JSA mechanism to address the native demand for multi-scale learning of user interaction sequences on top of hybrid tokenization. 
Inspired by the recent development of the Native Sparse Attention (NSA) mechanism from DeepSeek-AI \cite{yuan2502native}, we design a gated sparse attention strategy as follows: 
\begin{equation}
    \textbf{o} = \sum g_j \cdot \textbf{SparseAttn}_j(Q, K, V)
    \label{eq:gated_output}
\end{equation}
\noindent where $Q, K, V \in \mathbb{R}^{|dm \times L|}$, $dm$ is the hidden dimension and $L$ is the sequence length. The sparse attention strategy is the same for all the heads. Thus, we omit the multi-head expression in the following sections for simplicity. 
We propose four sparse attention strategies corresponding to different modeling requirements of user interaction sequences. 

\subsubsection{Multi-journey Compression}
A shopping journey usually refers to a behavior sequence with constant user interest on an e-commerce platform.
A user interaction sequence could cover multiple underlying shopping journeys. For example, a husband could shop for gifts for his wife while browsing electronic items such as a desktop computer for his workstation. Understanding the multi-journey nature is important to model the current journey comprehensively. We segment the token sequence into blocks size $l$ and a stride size $d \leq l$. We apply a Multi-layer Perception (MLP) layer on each block to compress the information of the key and value parameters accordingly,
\begin{equation}
    \bar{K}_{block} = \textbf{MLP}(K_{block}), \text{ } \bar{V}_{block} = \textbf{MLP}(V_{block}),
\end{equation}
\noindent where $K_{block}, V_{block} \in \mathbb{R}^{|dm \times l|}$ and $\bar{K}_{block}, \bar{V}_{block} \in \mathbb{R}^{|dm \times 1|}$. Then all the compressed key blocks are concatenated to get
\begin{equation}
    \widetilde{K}_{comp} = \textbf{Concat}([\bar{K}_{block}^1, ..., \bar{K}_{block}^{\lfloor{\frac{L-l}{d}}\rfloor}]),
\end{equation} and vice versa for value blocks. We finally compute the attention by the following equation to get the compressed output information: 
\begin{equation}
    \textbf{o}^{comp} = \textbf{Attn}(Q, \widetilde{K}_{comp}, \widetilde{V}_{comp})
    \label{eq:sparse_attn_comp}
\end{equation}

\subsubsection{Intra-journey Selection}
Intra-journey is what the customer is continuously viewing that is linked with each other, e.g., soccer balls and soccer shirts for a soccer season. However, they might not be adjacent due to other journeys interleaved.  
To better model the intra-journey item transition, we develop the second sparse attention strategy. We consider the top-$N$ most important blocks and concatenate them to form a new sequence for attention learning. 
We consider the compressed block-level knowledge in the first strategy based on the query and key, and compute the importance weight over blocks for top-$N$ selection, 
\begin{equation}
    \pi_{select} = \textbf{Softmax}(\textbf{Sum}(Q^T \widetilde{K}_{comp}, \text{axis=0})),
\end{equation} where $\pi_{select} \in \mathbf{R}^{|1 \times \lfloor{\frac{L-l}{d}}\rfloor|}$. We select top-$N$ blocks with the highest importance score and concatenate the selected (key, value) to form the new (key, value) for attention computation: 
\label{eq:selected_block}
   \begin{align}
    \widetilde{K}_{select} &= \textbf{Concat}([K_{block},... ]_{selected}) \\
    \widetilde{V}_{select} &= \textbf{Concat}([V_{block},... ]_{selected})
\end{align} 
\begin{equation}
    \textbf{o}^{intra} = \textbf{Attn}(Q, \widetilde{K}_{select}, \widetilde{V}_{select})
    \label{eq:sparse_attn_intra}
\end{equation}

\begin{figure*}[t!]
    \centering
    \includegraphics[width=0.8\linewidth]{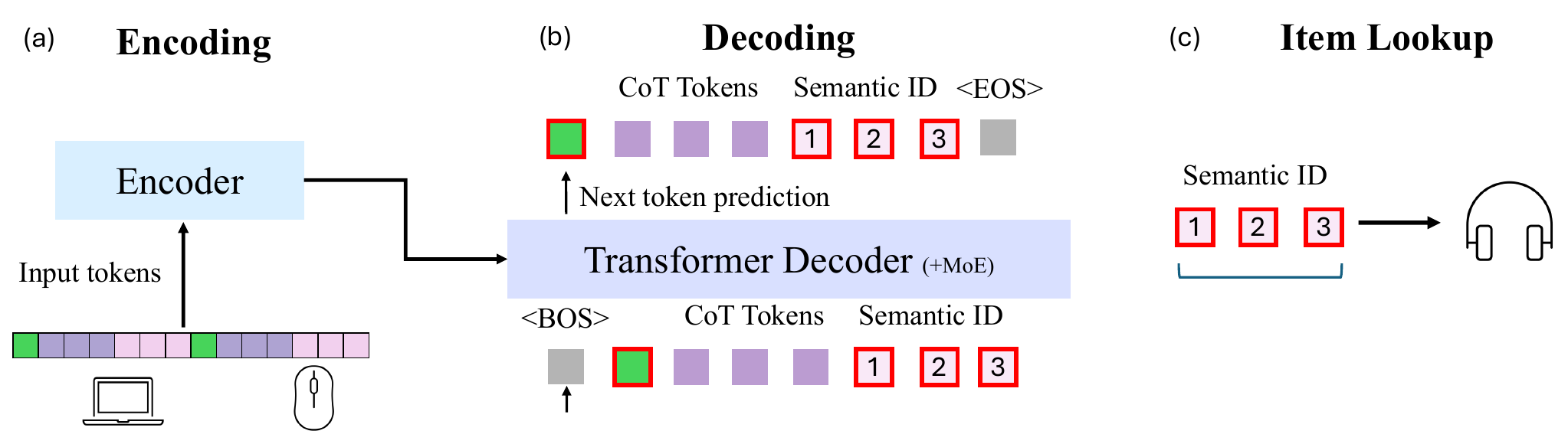}
    \caption{Inference of GRACE.
    }
    \label{nsainfer}
\end{figure*}

\subsubsection{Inter-journey Transition}
Inter-journey is the transition between intra-journeys, e.g., the customer first views soccer balls and later TV tables.
Inter-journey transition becomes a useful context when we focus on the high-level journey transition and correlation without getting too many details at the item-level, and learn more generic shopping patterns under the multi-behavior setting. To construct the high-level traversal of the journey transition over user interaction sequence, we keep $M_g$ graph CoT tokens and $M_s$ semantic tokens per item at the desired level and concatenate them to form a new sequence for inter-journey learning. 
For example, we set $M_g = M_s = 1$ in Figure \ref{figure2} and keep the first tokens from CoT and semantic tokens per item based on the position. We drop unselected tokens and build new key and value, $\widetilde{K}_{tokens}, \widetilde{V}_{tokens}$ respectively, by concatenating the selected tokens. The output of this attention is by, 
\begin{equation}
    \textbf{o}^{inter} = \textbf{Attn}(Q, \widetilde{K}_{tokens}, \widetilde{V}_{tokens})
    \label{eq:sparse_attn_inter}
\end{equation}

\subsubsection{Current-journey Comprehension}
Finally, the user might follow a momentum on the current shopping journey, which leads to the design of the last sparse attention strategy with local attention. 
We apply a truncated window with size $w$ on the input sequence to select the most recent $w$ (key, value) as $\widetilde{K}_{current}, \widetilde{V}_{current} \in \mathbb{R}^{|dm \times w|}$ for attention computation, 
\begin{equation}
    \textbf{o}^{current} = \textbf{Attn}(Q^T, \widetilde{K}_{current}, \widetilde{V}_{current})
\end{equation}

\subsubsection{Gated Output of Sparse Attention}
After we apply four different sparse attention mechanisms, we get four outputs: $\textbf{o}^{comp}$, $\textbf{o}^{intra}$, $\textbf{o}^{inter}$, $\textbf{o}^{current}$. We aggregate these outputs through a weighted average by Eq.\ref{eq:gated_output}. The gate weight $g_j$ is updated during the training. 
The final aggregated output $\textbf{o}$ could be parsed by following layers of the attention blocks.

\subsection{Objective Function and Inference}
Our sparse attention design could be generalized based on the underlying transformer models. Thus, the training objective functions are related to the selected transformer model. Despite various transformer setups, a common objective is the causal language modeling loss based on the predicted output and labels. Cross-entropy loss is used to measure the gap between prediction and labels, 
\begin{equation}
    \mathcal{L}_{ce} = -\sum_{batch} y\cdot\log(P(b_n, v_n | S_u, \Phi, \Psi)),
\end{equation} where $y$ is the label and $\Phi, \Psi$ refer to all learnable parameters in the encoder and the decoder, respectively. Note that the tokenization is based on the pretrained embeddings and the existing PKG, without involving trainable parameters.

At the inference stage, the encoder takes the input token sequence and extracts information for the decoder to predict items. Typically, we start the item prediction with the `<BOS>' token at the decoder and predict the behavior token first. After that, we generate all the CoT tokens for better guided predictions on the item semantic tokens, which will be used to identify the item. We conduct beam-search style token retrieval at each position to generate possible item candidates, with the number of beams $n_{beam}$ as a configurable parameter. 
Based on all the retrieved item candidates, we use the probability score of $P(b_n, v_n | s_u)$ for item ranking, which could be further used in evaluation and downstream applications. We summarize the inference process in Figure \ref{nsainfer}.

\begin{table}[H]
    \centering
    \caption{Data Description. }
    \label{tab:data_stats}
    \begin{tabular}{lcc}
    \hline 
         & \textbf{Home} & \textbf{Electronics} \\
         \hline 
         \#Items & 48,421 & 32,098\\ 
         \#Item/\#User Ratio & 4.84  & 3.21\\
         \hline 
         Avg. \#Seq. Length & 171.54 & 85.43\\
         Avg. \#ATC & 57.54 (33.54\%)  & 18.33 (21.46\%) \\
         Avg. \#Click & 171.54 (54.89\%) & 58.75 (68.77\%) \\
         Avg. \#Like & 2.31 (1.35\%) & 1.39 (1.63\%)\\
          Avg. \#Remove & 17.53 (10.22\%) & 6.96 (8.15\%)\\
        
         \hline
    \end{tabular}
    \label{dataset}
\end{table}


\section{EXPERIMENTS}
\subsection{Experiment Setup}
\subsubsection{Datasets} To evaluate the effectiveness of CoT tokens and JSA, we collected a real-world customer behavior dataset from Walmart.com, one of the large online shopping platforms in the US, for two categories, Home and Electronics. Click, add-to-cart (ATC), like, and remove-from-cart (Remove) behaviors are collected in our datasets.
The raw multi-behavior user-item interactions in the real world could encourage trivial solutions due to the connection between behaviors. For example, an ATC of an item is usually a consequence of a view of the same item, leading to a same-item prediction.
To avoid learning trivial solutions, we further process the data by merging the items with multiple associated behaviors and keeping the prioritized behavior based on the order, ATC, Like, and then Click. In this case, if an item is clicked and then added to cart, we only keep the add-to-cart behavior, forcing the model to learn the sequence transformation and predict new items and behaviors.
Table \ref{dataset} summarizes the details of our datasets. Each dataset contains 10000 user sequences. 
Please note that the prediction accuracy of our datasets is around 20\% maximum because of the harder dataset after applying the aforementioned `behavior merging' strategy.
Our datasets also have a larger $\#Item/\#User$ ratio and are sparser compared with the publicly available datasets \cite{liu2024multi}, which also makes the item prediction target more challenging.

\begin{table*}[h!]
\caption{Comparison with generative recommendation models for target behavior prediction (* p-value < 0.01).
}
\begin{tabular}{>{\centering\arraybackslash}p{2.5cm}ccccccccc}
\toprule
\multirow{2}{*}{\textbf{Model Type}} & \multirow{2}{*}{\textbf{Model}} & \multicolumn{4}{c}{\textbf{Home}}   & \multicolumn{4}{c}{\textbf{Electronic}}\\
\cmidrule(r){3-6} \cmidrule(r){7-10} 
                                         &                         & HR@5 & NDCG@5 & HR@10 & NDCG@10 & HR@5 & NDCG@5 & HR@10 & NDCG@10 \\
 
\midrule
\multirow{4}{*}{\shortstack{Sequential \\ Recommendation}}  & GRU4Rec                 & 1.50 & 1.09   & 1.75  & 1.17    &    6.48 &    5.11    &   7.62   &   5.49      \\
                                         & BERT4Rec$_{\textit{M}}$ & 1.32 & 1.04   & 2.06  & 1.27    &   18.02   &    13.35   &  23.57     &     15.13    \\
                                         & SASRec$_{\textit{M}}$   & 5.13 & 3.58   & 6.36  & 3.98    &   18.94   &    14.10    &  21.94     &    15.07     \\
                                         & RETR   & 5.57 & 3.68   & 7.31  & 4.63    &   21.77   &    15.93    &  24.22     &    17.49    \\
                                        & TIGER   & 2.70 & 1.80   & 4.07  & 2.24    &   14.97   &    10.93    &     19.60  &      12.42   \\
\midrule
\multirow{3}{*}{\shortstack{Multi-Behavior\\Sequential \\Recommendation}}  & BERT4Rec$_{\textit{B}}$ & 1.44 & 1.10   & 1.90  & 1.24    &   7.02   &   5.13     &  9.85     &     6.05    \\
                                         & SASRec$_{\textit{B}}$   & 1.93 & 1.46   & 2.36  & 1.60    &  6.91    &   5.03     &    8.44   &   5.52      \\
                                         & MBHT     &  2.26          & 1.69 & 2.97   & 1.84 &  19.77   &    15.67  &     23.95   &       16.84   \\   
                                         & MB-STR                  & 2.12 & 1.51   & 2.84  & 1.73    &    18.97  &   14.48     &    22.23   & 15.41  \\
                                         
\midrule                                         
\multirow{3}{*}{\shortstack{Multi-Behavior\\Generative \\Recommendation}}                                           & MBGen          & 5.88 & 3.95   & 8.52  & 4.80    & 23.16   & 16.56      & 27.83      & 18.00     \\

                                         & \textbf{GRACE}           & \textbf{11.73}* & \textbf{8.00}*   & \textbf{17.63}* & \textbf{9.92}*    &    \textbf{26.40}* & \textbf{19.17}*   & \textbf{33.97}* & \textbf{21.49}*         \\
                                        & \textbf{\#Improve}   
                                        & +99.48\%
                                        & +102.53\%
                                        & +106.92\%
                                        & +106.67\%
                                        & +13.99\%
                                        & +15.76\%
                                        & +22.06\%
                                        & +19.39\%\\
                                         
\bottomrule                                         
\end{tabular}
    \label{mainresult}

\end{table*}

\subsubsection{Baselines}
We compare GRACE with 8 state-of-the-art recommendation models:
\begin{list}{$\square$}{\leftmargin=1em \itemindent=0em}
  \item \textit{\textbf{Sequential Recommender Systems}}: We consider the first Recurrent Neural Networks (RNNs)-based sequential recommender
  \textbf{GRU4Rec}~\cite{hidasi2015session}, two Transformer-based baselines \textbf{SASRec$_{\textit{M}}$}~\cite{kang2018self} \textbf{BERT4Rec$_{\textit{M}}$}~\cite{sun2019bert4rec}, and one pathway-aware sequential recommender
  \textbf{RETR}~\cite{yao2024recommender}.
  We also compare our model with \textbf{TIGER}~\cite{rajput2023recommender} for generative sequential recommendation.

  \item \textit{\textbf{Multi-behavior Sequential Recommender Systems}}: 
  \textbf{MB-STR} learned heterogeneous item-level item information and generated behavior-awared predictions~\cite{yuan2022multi}. \textbf{MBHT}~\cite{yang2022multi} captured both short-term and long-term cross-type behavior dependencies with a hypergraph-enhanced transformer.
  We modified SASRec and BERT4Rec into
\textbf{SASRec$_{\textit{B}}$} and \textbf{BERT4Rec$_{\textit{B}}$} as two other baselines. With the subscript $_\textit{B}$, an item will be assigned with different embeddings when the associated behaviors are different, increasing the item embeddings size from $|\mathcal{V}|$ to $|\mathcal{V}| \times |\mathcal{B}|$.
  We also compare our model with \textbf{MBGen}~\cite{liu2024multi} for multi-behavior generative sequential recommendation.

\end{list}

\subsubsection{Implementation Details}
For item semantic tokenization, we generate 128-dimensional item embeddings based on item textual information such as titles and descriptions through a pre-trained BERT model.
We adopt the balanced semantic ID approach using RQ-VAE only for the first-level tokens. At level-1, the hidden dimension in the encoder of RQ-VAE is 2048, and the decoder is equipped with a ReLU activation function. The codebook size is 64, and the hidden dimension is 32. 
At level-2, the residual vectors of items with the same level-1 token are clustered using K-means clustering with $K=64$. The cluster index is the level-2 code. A random token is assigned as the level-3 token to avoid collisions \cite{liu2024multi}.
For item CoT tokenization, we leverage the product knowledge graph and extract product types (PT), item price (PRICE), and brand (BRAND) information. Typically, we tokenize item price by categorizing the price value into five price bands (PB). We construct CoT tokens based on the scope of the information in descending order to imitate the item navigation process, [$PT\_token$, $PRICE\_token$, $BRAND\_token$]. 
We choose the Switch Transformers \cite{fedus2022switch} as the backbone of our GRACE. The hidden dimension of the model is 256, with 512 dimensions in the ReLU activation function, and 6 heads of dimension 64 in the self-attention layer. 
We set the batch size to 512 with a learning rate of 0.001 and trained our model with the AdamW optimizer for 50000 steps on an A100 GPU.

\subsubsection{Evaluation Metrics}
We adopt two widely-used metric, Recall@K and NDCG@K, where $K \in \{5, 10\}$. 
For dataset splitting, we apply the leave-one-out strategy used in \cite{liu2024multi}, and the ground-truth item of each sequence is ranked among all other items instead of some sampled negative items for reliable evaluation. We conduct a beam search with 10 beams to rank the top 10 candidates in all items to evaluate the Recall@K and NDCG@K metrics for the proposed GRACE model. All item sequences are truncated to 50 for training and evaluating our proposed model and the baselines. We report metrics under (1) target behavior (ATC) item prediction task, (2) behavior-specific item prediction task, and (3) behavior-item prediction task.

\subsection{Overall Performance}

Table \ref{mainresult} presents the performance of eight baseline recommendation models alongside our proposed GRACE across the Home and Electronic datasets. We report two standard ranking metrics, Recall (HR@K) and Normalized Discounted Cumulative Gain (NDCG@K) with K=5 and 10, to evaluate both the accuracy and ranking quality.

Traditional sequential models such as GRU4Rec, BERT4Rec\textsubscript{M}, SASRec\textsubscript{M} demonstrate modest performance. With pathway attention, RETR achieves the strongest scores in this group (HR@10 of 7.31 on Home and 24.22 on Electronic), indicating the need for user shopping trajectory modeling. However, these models operate on flat item sequences and are limited in capturing multi-behavioral context or hierarchical user intent, leading to suboptimal generalization in real-world settings.
TIGER shows moderate improvements, particularly on the Electronic dataset. It falls short of capturing the nuanced, multi-level patterns inherent in complex user journeys.

SASRec$_{\textit{B}}$ and BERT4Rec$_{\textit{B}}$ show lower performance than their sequential recommendation variants, indicating the limited behavior modeling capability.
Multi-behavior models like MBHT and MB-STR better accommodate behavior diversity, with MBHT reaching HR@10 of 23.95 on Electronic. These models integrate behavior signals more directly but still lack the generative capacity to model token-level transitions and long-range dependencies effectively, especially on the Home dataset with a high $\#Item/\#User$ ratio and more sparsity after behavior merging.  

MBGen, the strongest baseline in the multi-behavior generative category, significantly boosts performance by incorporating behavioral semantics into the generative process (HR@10 of 8.52 on Home and 27.83 on Electronic). However, it relies on static tokenization and full attention, which limits its scalability and expressiveness in sparse or behavior-rich environments.

Our proposed model, GRACE, outperforms all baselines across every metric and dataset. On the Home dataset, GRACE achieves HR@10 of 17.63 and NDCG@10 of 9.92, representing relative improvements of +106.92\% and +106.67\% over MBGen. On the Electronic dataset, GRACE attains HR@10 of 33.97 and NDCG@10 of 21.49, yielding +22.06\% and +19.39\% gains, respectively. {\color{black} The larger gain in the Home is due to greater diversity (higher \#Item/\#User Ratio in Table-1) in the product catalog and relatively lower baseline performance on the complex behaviors~\cite{kang2018self, yang2022multi}, which CoT tokenization and JSA leverage more effectively. In contrast, Electronics has more homogeneous items and shorter interaction chains, showing relatively easier tasks.}

{\color{black} The performance variance of multi-behavior, sequential, and generative recommenders is also related to the complexity of customer behavior types and the catalog size. Sequential recommenders focus more on item-transition, while multi-behavior recommenders cannot balance the learning of behavior and item-transition easily when the \#Item/\#User Ratio is high and negative behaviors (remove-from-cart) exist. Multi-behavior generative recommenders perform better due to the richer item representation and better behavior-item reasoning.
By leveraging hybrid token structures that capture semantic, behavioral, and graph-based CoT signals, GRACE forms richer input representations. JSA mechanism—tailored to compressed, intra-, inter-, and current-journey signals—enables efficient and expressive learning across long sequences. As a result, GRACE delivers both high accuracy and scalability, positioning it as a practical and effective solution for next-generation multi-behavior generative recommendation systems. 
}

\begin{table*}[h!]
\centering
\caption{Results of ablation studies for three multi-behavior tasks on the Home dataset. 
}
\begin{tabular}{p{5cm}cccccc}
\hline
\multicolumn{1}{c}{\multirow{3}{*}{\textbf{Model}}} 
 &
  \multicolumn{2}{c}{\textbf{Target Behavior}} &
  \multicolumn{2}{c}{\textbf{Behavior Specific}} &
  \multicolumn{2}{c}{\textbf{Behavior-Item}} 
  \\ 
\cmidrule(r){2-3} \cmidrule(r){4-5} \cmidrule(r){6-7}  
\multicolumn{1}{c}{}                                & NDCG@5 & NDCG@10 & NDCG@5 & NDCG@10 & NDCG@5 & NDCG@10 
\\ \hline
\multicolumn{7}{c}{\textbf{CoT Tokenization Ablation w/o JSA }} \\ \hline
w/o JSA&
 6.96 &
 8.79 &
 8.65 &
 10.42 &
 4.95 &
 6.13 
  \\
w/o BRAND \& JSA &
 \underline{7.17} &
 8.72 &
 10.04 &
 11.70 &
 5.62 &
 6.91 
  \\
w/o PRICE \& BRAND \& JSA&
  6.93 &
  8.18 &
  9.39 &
  10.99 &
  5.32 &
  6.54 
  \\
w/o CoT \& JSA (i.e., MBGen)&
  3.95 &
  4.80 &
  9.83 &
  10.97 &
  6.45 &
  7.49
  \\
  \hline
\multicolumn{7}{c}{\textbf{CoT Tokenization Ablation w/ JSA }} \\ \hline
w/o BRAND &
 7.05 &
 \underline{8.82} &
 10.55 &
 12.30 &
 6.06 &
 7.36 
  \\
w/o PRICE \& BRAND &
  6.50 &
  7.96 &
  \underline{10.86} &
  \underline{12.48} &
  6.06 &
  7.44 
  \\
w/o CoT (i.e.,  MBGen+JSA)&
 4.00 &
 4.84 &
 10.70 &
 11.88 &
 \textbf{7.03} &
 \textbf{8.16} 
  \\
 \hline
\multicolumn{7}{c}{\textbf{Journal-Aware Attention Ablation w/o CoT Tokens}} \\ \hline
w/o CoT \& Intra. &
  2.90 &
  3.55 &
  8.81 &
  9.79 &
  6.10 &
  6.87 
   \\
w/o CoT \& Comp. \& Intra.  &
  2.47 &
  3.16 &
  8.43 &
  9.42 &
  4.78 &
  5.50 
   \\
w/o CoT \& Inter. &
  2.40 &
  2.92 &
  7.67 &
  8.63 &
  5.09 &
  5.88 
   \\
w/o CoT \& Truncated Window &
 2.97 &
 3.79 &
 8.81 &
 9.76 &
 5.82 &
 6.67 
  \\
  \hline

GRACE &
 \textbf{8.00} &
\textbf{ 9.92} &
 \textbf{11.09} &
 \textbf{13.01} &
 \underline{6.73} &
 \underline{7.69} 
  \\  \hline
\end{tabular}%
\label{ablation}
\end{table*}

\subsection{Ablation Study}
\subsubsection{CoT Tokenization Ablation}
We remove CoT tokens (PT, PRICE, BRAND) one by one and study their impacts on NDCG of three tasks with and without JSA on the Home dataset (Table \ref{ablation}) respectively. 
Removing CoT tokens causes a lower target behavior NDCG, with the lowest scores achieved when all CoT tokens are removed in the variants "w/o CoT \& JSA" and "w/o CoT", leading to more than a 50\% drop in NDCG@10 compared with GRACE. Note that "w/o CoT \& JSA" and "w/o CoT" are equal to MBGen and MBGen+JSA, respectively. 
Without JSA, fewer CoT tokens lead to better NDCG scores in the behavior-specific and behavior-item tasks, which is against the trend of the target behavior task. However, the existence of JSA balances the target behavior prediction and the other two tasks well, and GRACE outperforms "w/o JSA" on all tasks.
This demonstrates that CoT tokens provide better target-oriented information on the target behavior prediction, Add-to-cart, with a strong business value due to the positive correlation between cart size increase and revenue growth. At the same time, more CoT tokens imply more complicated learning patterns, requiring a more optimized attention mechanism like JSA to balance multi-behavior learning.

\subsubsection{JSA Strategy Ablation}
To understand the effectiveness of each strategy in JSA, we conduct ablation studies by removing each strategy without CoT tokens (Table \ref{ablation}). 
We observe that removing any strategy reduces NDCG scores of all three tasks compared with GRACE, with the biggest drop on the target behavior tasks by removing the compression 
and the intra-journey selection (NDCG@10=3.16). It demonstrates the significance of compressing and selecting the user interaction segments when predicting the next Add-to-cart items. 
Removing the inter-journey transition significantly reduces NDCG scores of the behavior-specific (NDCG@10=7.30) and the behavior-item (NDCG@10=4.13) tasks. It supports our hypothesis that high-level journey transition and correlation are important to multi-behavior predictions, and directly applying sparse attention like NSA~\cite{yuan2025native} overlooks the complexity of user-item interactions.

\subsection{Hyper-parameter Analysis}
\begin{figure}
    \centering
    \includegraphics[width=\linewidth]{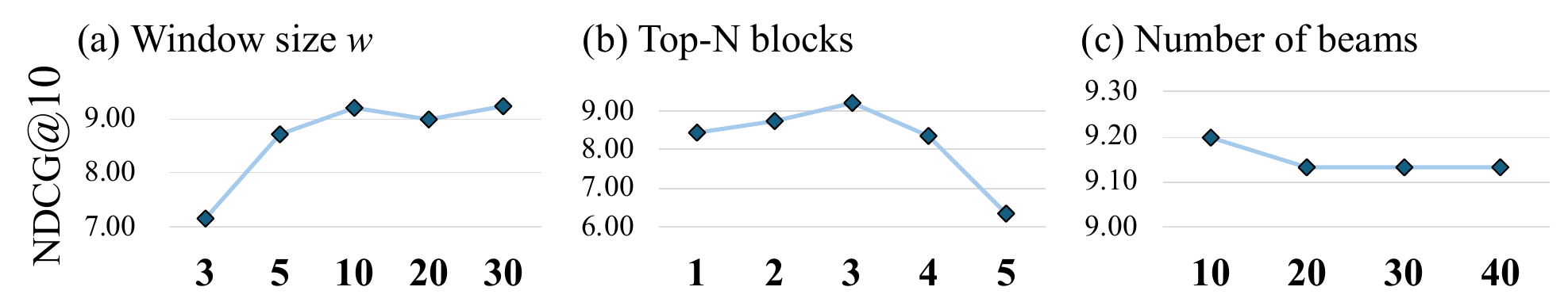}
    \caption{Hyper-parameter analysis}
    \label{parameter_analysis}
\end{figure}

We conduct a sensitivity study on key hyper-parameters in GRACE to assess their impact on model performance. Figure \ref{parameter_analysis} presents hyper-parameter analysis across variations in window size $w$ for current-journey, top-$N$ selection for intra-journey attention, and beam width used during inference. As shown in Figure 4(a), performance improves significantly when increasing $w$ from 3 to 10. Further increases to 20 yield diminishing returns, while setting $w$=30 slightly improves performance but adds more computation. Based on this trade-off, we set $w$=10 to balance efficiency and expressiveness. Figure 4(b) shows that using top-3 blocks in intra-journey attention achieves the best result, outperforming both lower (top-$N$=1) and higher (top-$N$=4, 5) values. Selecting too few blocks fails to capture sufficient context, while selecting a larger n may introduce irrelevant noise. We therefore fix top-$N$=3 in our final setup. During inference, we vary the beam size in the range of 10 to 40 as shown in Figure 4(c). The results indicate that increasing the beam size beyond 10 does not significantly improve performance. Thus, we use a beam size of 10. MBGen generally requires a larger number of beams, indicating the effectiveness of CoT tokens and JSA.  Overall, GRACE exhibits robust performance across a wide range of hyper-parameters, with optimal results obtained at $w$=10, top-$N$=3, and beam size=10.

\subsection{Computational Efficiency}

To evaluate the efficiency benefits of our journal-aware sparse attention mechanism, we compare GRACE’s attention layer against standard full attention under varying item sequence lengths. Table \ref{efficiency} reports the total number of activated parameters in both settings, highlighting how GRACE reduces computation by dynamically focusing on informative tokens. At a sequence length of 50, GRACE activates 43,092 parameters compared to 63,504 in full attention, resulting in a 32\% reduction. This efficiency gain becomes more significant as sequence length increases, with reductions of 43\% and 48\% at lengths 100 and 200, respectively. These improvements stem from the structured design of our attention mechanism, which decomposes user interaction sequences into intra-, inter-, and current journeys and applies targeted sparse attention over each.

\begin{table}[h]
\caption{Activated parameters in sparse attention}
\resizebox{\columnwidth}{!}{%
\begin{tabular}{cccc}
\hline
\textbf{Item Sequence Lengths}               & \textbf{50} & \textbf{100} & \textbf{200} \\ \hline
Full Attention            & 63,504      & 252,004      & 1,004,004    \\
GRACE Attention           & 43,092      & 144,576      & 522,042      \\ \hline
\textbf{Activated Parameters Reduced} & -32\%        & -43\%         & -48\%         \\ \hline

\end{tabular}%
}
    \label{efficiency}

\end{table}

\subsection{Multi-behavior Recommendation}

To further evaluate the effectiveness of GRACE in handling diverse user intents, we conduct a fine-grained comparison with MBGen on multi-behavior recommendation: Add-to-cart (ATC), Click, Like, and Remove-from-cart (Remove). Table~\ref{tab:mbgen-comparison} reports NDCG@10 for each behavior on both Home and Electronic datasets. GRACE consistently outperforms MBGen on most behaviors and domains. In the Home dataset, GRACE achieves notable improvements in ATC (+62.5\%) and Click (+16.8\%) behaviors, highlighting its ability to model high-intent and exploratory actions with finer context. In the Electronic domain, GRACE shows even greater gains, particularly in Like behavior, where it improves NDCG@10 from 9.00 to 21.22—more than doubling the effectiveness. These improvements suggest that GRACE’s hybrid tokenization and JSA design are especially effective for capturing nuanced and less frequent behaviors, which are often critical for downstream engagement and personalization.

\begin{table}[h]
\caption{Multi-behavior Recommendation NDCG@10}

    \centering
    \begin{tabular}{lcccc}
        \hline
        \multirow{2}{*}{\textbf{Behaviors}} & \multicolumn{2}{c}{\textbf{Home}} & \multicolumn{2}{c}{\textbf{Electronic}} \\ 
        \cline{2-5}
         & MBGen & GRACE (Our) & MBGen & GRACE (Our)  \\ 
        \hline
ATC      & 4.80      & 9.92     & 18.00      & 21.49       \\
Click            & 7.46      & 10.48     & 15.56       & 27.44      \\
Like  & 28.22      & 18.76     & 9.00      & 16.97      \\
Remove & 33.67      & 27.03     & 46.74       & 41.07      \\
        \hline
    \end{tabular}
\label{tab:mbgen-comparison}
\end{table}

\subsection{Validation with CoT Tokenization}

\begin{figure}[h]
    \centering
    \includegraphics[width=0.639\linewidth]{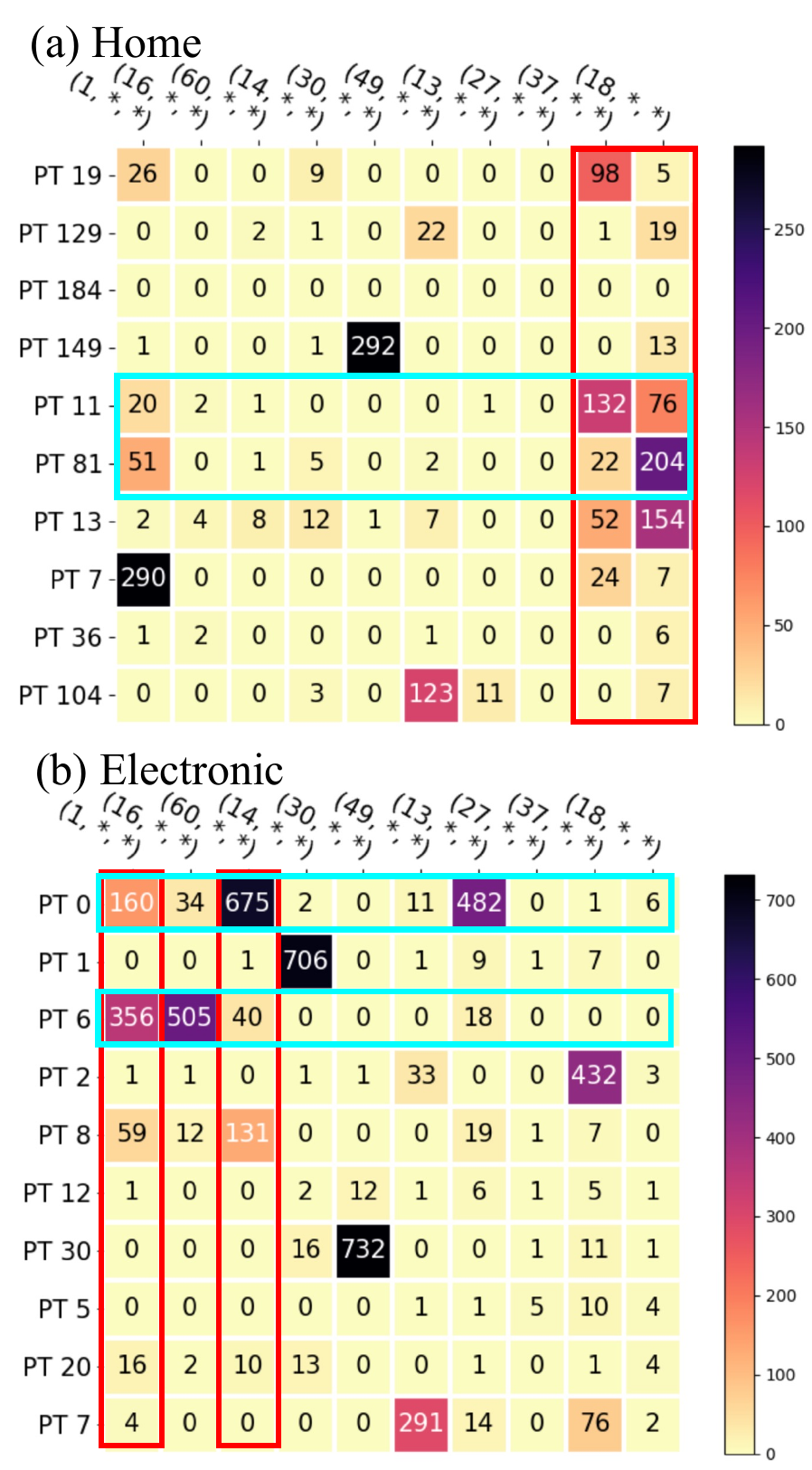}
    \caption{CoT-PT and L1 tokens co-occurrence heatmap. 
    }
    \label{TokenHeatMap}
\end{figure}

To assess the effectiveness of Chain-of-Thought tokenization in structuring the item space, we present a co-occurrence heatmap in Figure~\ref{TokenHeatMap}, showing the distribution of items across product type (PT) tokens and semantic token level-1 (L1) clusters for the Home and Electronic datasets. Each cell indicates the number of items that share the corresponding PT and L1 token pair.

The visualization highlights a key benefit of incorporating product type information explicitly into the tokenization pipeline: CoT tokenization significantly narrows the effective search space during item generation. For instance, in the Home dataset (Figure \ref{TokenHeatMap}a), PT tokens such as PT=11 and PT=81 are tightly aligned with specific semantic tokens (L1=37, 18), reducing token dispersion and creating highly concentrated activation zones. Only considering semantic tokens might be distracted by other different items with shared L1 semantic tokens, e.g., PT=19. This demonstrates that GRACE can focus decoding within localized semantic clusters once a product category is known. A similar pattern is also observed in the Electronic dataset (Figure \ref{TokenHeatMap}b). 
These sharply defined token regions illustrate how CoT tokenization acts as a semantic filter, effectively pruning the space of candidate tokens early in the generation process.

\section{Related Works}
\noindent{\textbf{Sequential Recommendation (SR)}}: SR predicts the next item of user interest based on user history~\cite{rendle2010factorizing,hidasi2015session,chang2021sequential,wu2019session,tang2018personalized, de2021transformers4rec}. 
It has been observed that the transformer model~\cite{vaswani2017attention} significantly improves performance in sequential recommendation tasks. SASRec~\cite{kang2018self} applied the Transformer decoder to model item sequences. BERT4Rec \cite{sun2019bert4rec} leveraged the bi-directional attention to predict masked items.
Yao et al. \cite{yao2024recommender} introduced the pathway attention mechanism through multiple small pathways in the network.

\noindent{\textbf{Multi-behavior Sequential Recommendation (MBSR) }}: MBSR problem predicts the items the user will interact with by incorporating behavior types of history user-item interactions~\cite{guo2019buying, gu2020deep, yuan2022multi, yang2022multi, xia2022multi, su2023personalized}.  
DIPN~\cite{guo2019buying} and DMT~\cite{gu2020deep} captured the sequential pattern with a fixed single behavior. MB-STR~\cite{yuan2022multi} learned heterogeneous item-level item information and generated behavior-aware predictions. MBHT~\cite{yang2022multi} captured both short-term and long-term cross-type behavior dependencies with a hypergraph-enhanced transformer.
TGT~\cite{xia2022multi} leveraged the temporal graph Transformer to model time-evolving user preference, and PBAT~\cite{su2023personalized} addressed the co-influence between behavior correlations and item collaborations to personalize item recommendation. 
 
\noindent{\textbf{Generative Recommendation (GR)}}: GR systems have increasingly adopted tokenization techniques to structure user interaction sequences, enabling efficient encoding of behavioral patterns.
Traditional sequential models~\cite{kang2018self,sun2019bert4rec,hou2022core,zhang2019feature,zhou2020s3} relied on raw item ID sequences for modeling user interactions. However, these methods often struggled to capture higher-level behavioral contexts and suffered from scalability issues. 
Recent advancements introduced tokenization strategies to mitigate these shortcomings \cite{rajput2023recommender, jin2023language, li2023gpt4rec, petrov2023generative, zhai2024actions, liu2024mmgrec, wang2024eager, xiao2025progressive, hou2025actionpiece, deng2025onerec}.
TIGER \cite{rajput2023recommender} was one of the first methods to apply tokenization on T5 sentence embedding~\cite{ni2021sentence} through RQ-VAE~\cite{zeghidour2021soundstream} based on item descriptive text.
MMGRec~\cite{li2023gpt4rec} tokenized the multimodal information of each item to support multimodal recommendation. 
EAGER~\cite{wang2024eager} explored the tokenization of both behaviors and items to enhance behavior-semantic collaboration. ActionPiece~\cite{hou2025actionpiece} contextually tokenized behavior-item sequences by context-dependent tokenization. 
MBGen \cite{liu2024multi} studied the balanced item semantic tokenization to improve multi-behavior generative recommendation.
OneRec~\cite{deng2025onerec} combined item tokenization and preference alignment to unify generative retrieval and ranking.

Despite the advances of generative recommendation models in MBSR, they 
ignore explicit knowledge like price and brand that are deterministic for item retrieval while searching for tokens. 
Without deterministic knowledge, generative recommenders cannot correctly reason the next item token because of implicit tokenization. 
Furthermore, tokenization increases the attention computation quadratically, causing inefficient learning of user journeys. 
The journey-aware sparse attention in GRACE can efficiently learn different levels of journey-awareness with less attention computations for better task orientation. 
GRACE dynamically learns sparse attention patterns tailored to user-item interactions, unlike fixed sparse mechanisms. This adaptability allows GRACE to focus computational resources efficiently on informative interactions, significantly reducing overhead while maintaining expressive capacity in generative recommendation settings.

\section{Conclusion}
We presented GRACE, a generative recommendation framework that combines CoT tokenization with journey-aware sparse attention to improve both accuracy and efficiency in multi-behavior sequential recommendation. By structuring user interactions through semantic and CoT tokens, GRACE captures high-level intent and fine-grained behavior patterns. 
Our journey-aware sparse attention mechanism further reduces computational overhead by dynamically attending to the most relevant tokens across compressed, intra-, inter-, and current journeys. 
Experiments on large-scale real-world datasets show that GRACE significantly outperforms existing baselines in both effectiveness and scalability.

\clearpage

\bibliographystyle{ACM-Reference-Format}
\bibliography{grace_recsys_reference}

\end{document}